\documentclass[journal]{IEEEtran}

\ifCLASSINFOpdf
\else
   \usepackage[dvips]{graphicx}
\fi
\usepackage{url}

\usepackage{cite}
\usepackage{verbatim}
\usepackage{amsmath,amssymb,amsfonts}
\usepackage{amsthm}
\usepackage{algorithm}
\usepackage{graphicx}
\usepackage{bm}
\usepackage{textcomp}
\usepackage{array}
\usepackage{xcolor}
\usepackage[colorlinks=true, allcolors=blue]{hyperref}
\usepackage[utf8]{inputenc}
\usepackage{algpseudocode}
\usepackage{soul}
\usepackage{multirow}
\usepackage{multicol}
\usepackage{booktabs}

\usepackage[caption=false,font=footnotesize]{subfig}

\newcommand{\ours}{CoMFed}

\theoremstyle{remark}

\begin{document}

\title{Communication-Efficient and Robust Multi-Modal Federated Learning via Latent-Space Consensus}

\author{Mohamed Badi, Chaouki Ben Issaid, and Mehdi Bennis%
\thanks{M. Badi, C. Ben Issaid, and M. Bennis are with the Center for Wireless Communications, University of Oulu, Oulu 90014, Finland (e-mail: \{mohamed.badi, chaouki.benissaid, mehdi.bennis\}@oulu.fi). This work was supported by the ERA-NET CHIST-ERA project MUSE-COM.}}

\maketitle

\begin{abstract}
Federated learning (FL) enables collaborative model training across distributed devices without sharing raw data, but applying FL to \emph{multi-modal} settings introduces significant challenges. Clients typically possess heterogeneous modalities and model architectures, making it difficult to align feature spaces efficiently while preserving privacy and minimizing communication costs. To address this, we introduce \emph{\ours{}}, a \textbf{Communication-Efficient Multi-Modal Federated Learning} framework that uses learnable projection matrices to generate compressed latent representations. A latent-space regularizer aligns these representations across clients, improving cross-modal consistency and robustness to outliers. Experiments on human activity recognition benchmarks show that \ours{} achieves competitive accuracy with minimal overhead.

\end{abstract}

\vspace{-0.3cm}
\section{Introduction}\label{section1}
Federated Learning (FL) \cite{mcmahan2017communication} enables distributed devices to collaboratively train machine learning models without sharing raw data, thereby preserving privacy and reducing reliance on centralized storage. While classical FL has been effective in settings with homogeneous data and consistent model architectures, real-world deployments rarely satisfy these assumptions, particularly in sensing-driven and edge deployments. Instead, data are often multi-modal—arising from sources such as images, audio, LiDAR, inertial sensors, and mmWave radar—and distributed across devices with distinct sensing capabilities and computational constraints. Enabling collaboration under such heterogeneous, multi-modal conditions introduces several fundamental challenges.

A central difficulty is communication efficiency. Exchanging model updates or gradients is already costly in standard FL, and this burden is amplified in multi-modal learning, where models are typically larger and architecturally diverse. Although numerous communication-reduction strategies have been developed—such as quantization, sparsification, scheduling, and second-order acceleration methods \cite{elgabli2024quantized,konevcny2016federated}—these approaches implicitly assume that clients exchange compatible gradient tensors. This assumption breaks down in multi-modal settings: clients may operate on different modalities, use different architectures, or even have different input dimensionalities, rendering gradient- or parameter-level averaging ill-defined.

To address multi-modal heterogeneity, \cite{ouyang2023harmony} proposed separating training across modality-combination subsets, such that only clients sharing identical modality availability collaborate using standard homogeneous FL schemes. While effective in structured settings, this partitioning limits cross-modal knowledge transfer and becomes restrictive when modality availability is sparse or highly fragmented.

Representation-based FL has therefore emerged as an alternative paradigm, in which clients exchange intermediate representations rather than full model parameters. Existing approaches, including distillation-based methods such as FedMD \cite{li2019fedmd}, typically rely on logits or features evaluated on a shared public dataset to align heterogeneous models. This reliance on public data and multi-stage training procedures significantly limits applicability in privacy-sensitive or resource-constrained deployments, where shared datasets may not exist.

{More broadly, prior work on multi-task and multi-modal FL considers heterogeneity in labels, inputs, or architectures \cite{issaid2025tackling,smith2017federated,che2023multimodal}, but does not directly address how to enable principled collaboration across architecturally diverse, multi-modal clients \emph{without} shared data, parameter alignment, or public supervision. Establishing a lightweight yet semantically meaningful coordination mechanism under these constraints remains an open challenge.}

Motivated by these challenges, {we propose a communication-efficient, architecture-agnostic,} multi-modal {FL} framework that enables collaboration across heterogeneous clients without requiring shared data, shared architectures, or homogeneous modality availability. {Unlike distillation-based or prototype-sharing approaches, the proposed method exchanges only \emph{class-wise low-dimensional latent statistics} computed on private data, avoiding public datasets and operating in a single-stage training process.} Our key idea is to learn client-specific projection matrices that map intermediate features into a shared latent space, enabling heterogeneous models to exchange semantically comparable information with minimal overhead. A robust alignment regularizer, based on geometric-median consensus, ensures consistency across clients while mitigating the impact of outliers and Byzantine behavior. {Because collaboration occurs entirely in a fixed-dimensional latent space independent of model size, the framework is inherently scalable to large client populations.}

Our contributions are summarized as follows:
\begin{itemize}
\item We propose a {latent-consensus} multi-modal FL framework that enables principled collaboration among clients with heterogeneous modalities and architectures {over arbitrary communication topologies}.
\item We develop a robust latent-space consensus mechanism leveraging geometric-median regularization to enhance resilience against outliers and Byzantine clients.
\item We demonstrate the effectiveness of the proposed method through comparisons with existing approaches on two real-world collected multi-modal datasets in FL settings.
\end{itemize}

Section \ref{sec:system} describes the symbols and notations used throughout the paper. Section \ref{sec:algorithm} details our framework and algorithms for both decentralized and centralized FL settings. Section \ref{sec:num} presents experimental results, and Section \ref{sec:conclusion} concludes the paper with final remarks.

\newcommand{\method}{CoMFed}


\vspace{-0.2cm}
\section{System Model and Problem Formulation}
\label{sec:system}

We consider a set of $N$ clients indexed by $i\in\{1,\dots,N\}$ connected through an undirected communication graph $G=(\mathcal{V},\mathcal{E})$ with neighbor sets $\mathcal{N}_i$.  
Each client $i$ holds a private dataset $\mathcal{D}_i=\bigcup_{m\in\mathcal{M}}\mathcal{D}_{i,m}$ drawn from a common label set $\mathcal{M}$ (not necessarily balanced across clients).  
Let $\mathcal{S}$ denote the set of available data modalities (e.g., vision, audio, or inertial sensors), and let $\mathcal{S}_i\subseteq \mathcal{S}$ represent the subset available to client $i$.  
Clients, therefore, possess different architectures suited to their modality subset and local resources.  
{At the network level, we assume synchronous and reliable communication over $G$, and abstract away physical-layer effects.}

Each client trains a local model $f_i(\mathbf{w}_i;\cdot)$ with parameters $\mathbf{w}_i$ by minimizing a private loss function
\begin{align}
\mathcal{L}_{\text{local}}^i(\mathbf{w}_i) = \mathbb{E}_{\xi\in\mathcal{D}_i}\big[\ell(f_i(\mathbf{w}_i;\xi),y)\big],
\end{align}
while simultaneously aiming to collaborate with other clients to improve generalization and robustness.  
The challenge is how to enable this collaboration efficiently and meaningfully when the models and inputs differ across clients.

\textbf{Why classical gradient averaging fails.}
{Traditional FL} approaches such as FedAvg~\cite{mcmahan2017communication} rely on gradient or weight averaging across clients.  
This requires that model parameters correspond semantically, which presumes homogeneous architectures and aligned feature spaces.  
In multi-modal FL, this assumption no longer holds: clients may have different input dimensions, modality subsets ($\mathcal{S}_i\neq \mathcal{S}_j$), and subsequently heterogeneous network structures, leading to incompatible parameterizations.  
Even when architectures are identical, different initializations can yield latent spaces with distinct geometries, making gradient averaging uninformative or even harmful.  
Hence, a different form of collaboration is needed—one that does not rely on direct parameter averaging.

\textbf{Toward representation-level collaboration.}
To uncover a meaningful common ground, we observe that while clients differ in their inputs and architectures, they share the same label space $\mathcal{M}$.  
Each local network can be interpreted as a nested composition of nonlinear transformations, and intermediate layers capture more abstract representations of the input successively.  
If we denote the feature output of the $l_i$-th layer of client $i$ by
\begin{align}
\mathbf{v}_i(\mathbf{w}_i^{l_i};\xi) = f_i^{l_i}\big(f_i^{l_i-1}(\dots f_i^{1}(\xi)\dots)\big) \in \mathbb{R}^{d_i},
\end{align}
then this feature lies in a client-specific representation space $\mathbb{R}^{d_i}$ that depends on the local architecture.

However, the dimensions $d_i$ and the semantic structure of these representations generally differ across clients.  
This misalignment prevents direct comparison or aggregation of features between clients—even if they correspond to the same label.  
To overcome this, we introduce a linear projection matrix $\mathbf{P}_i \in \mathbb{R}^{d \times d_i}$ as the simplest class of mappings from local features to a shared latent space $\mathbb{R}^d$ ($d \ll d_i$). Each $\mathbf{P}_i$ serves as a \emph{translator} that enforces a consistent, low-dimensional representation across clients, acting as an inductive bias and guiding them to learn how to communicate in a common latent language.

Formally, for any sample $\xi$ of class $m$, the projected feature is
\begin{align}
\mathbf{u}_{i,m} = \mathbf{P}_i\,\mathbf{v}_i(\mathbf{w}_i^{l_i};\xi)\in\mathbb{R}^{d}.
\end{align}

This latent-space design addresses the limitations of gradient-based aggregation: rather than averaging incompatible parameters, clients exchange information through compressed, semantically aligned representations that can be compared meaningfully.
By enforcing that representations corresponding to the same label map to a common region in the latent space, collaboration becomes possible despite architectural diversity and modality heterogeneity.

\textbf{Latent statistics and communication.}
To reduce communication cost, each client shares not all sample features but their per-class mean latent representations.  
For class $m$, we define
\begin{align}
\bar{\mathbf{v}}_{i,m} = \mathbb{E}_{\xi\in\mathcal{D}_{i,m}}\!\left[\mathbf{v}_i(\mathbf{w}_i^{l_i};\xi)\right],
\qquad
\mathbf{u}_{i,m} = \mathbf{P}_i \bar{\mathbf{v}}_{i,m}.
\end{align}
At round $t$, these are estimated from a minibatch $\tilde{\xi}_{i,m}^{(t)}$ as
\begin{align}
\widehat{\bar{\mathbf{v}}}_{i,m}^{(t)} = \frac{1}{|\tilde{\xi}_{i,m}^{(t)}|}\sum_{\xi\in\tilde{\xi}_{i,m}^{(t)}} \mathbf{v}_i(\mathbf{w}_i^{l_i,(t)};\xi),
\quad
\widehat{\mathbf{u}}_{i,m}^{(t)} = \mathbf{P}_i^{(t)}\,\widehat{\bar{\mathbf{v}}}_{i,m}^{(t)}.
\end{align}
Each client transmits the set $\{\widehat{\mathbf{u}}_{i,m}^{(t)}\}$ to its neighbors.

Given these translated representations, the goal is to encourage alignment among clients so that their latent embeddings for the same class become consistent across the network.  
To quantify the discrepancy between representations, we introduce a distance function $\phi(\cdot,\cdot)$ defined in the latent space.  
A well-designed $\phi$ should be convex and differentiable (or sub-differentiable), robust to outliers or noisy latent estimates (e.g., due to few samples), and tolerant to missing classes.  
This distance measure forms the basis of a \emph{latent consensus regularizer} that promotes representation alignment across connected clients.

For each client $i$, we define the latent-space regularizer as
\begin{align}
\mathcal{L}_{\text{reg}}^i(\mathbf{P}_i,\{\mathbf{u}_j\}_{j\in\mathcal{N}_i})
= \frac{1}{|\mathcal{N}_i|}\sum_{j\in\mathcal{N}_i}\phi\!\left(\mathbf{P}_i\,\mathbf{v}_i(\mathbf{w}_i^{l_i}),\ \mathbf{u}_{j}\right),
\label{eq:reg_term}
\end{align}
which measures how well the projected representations of client $i$ align with those of its neighbors $\mathcal{N}_i$ on the communication graph $G$.

The overall decentralized optimization objective then jointly minimizes each client’s local task loss and its regularization term
\begin{align}
\min_{\{\mathbf{w}_i,\mathbf{P}_i\}}\ 
\sum_{i=1}^{N}\Big(
\mathcal{L}_{\text{local}}^i(\mathbf{w}_i)
+ \lambda\,\mathcal{L}_{\text{reg}}^i(\mathbf{P}_i,\{\mathbf{u}_j\}_{j\in\mathcal{N}_i})
\Big).
\label{eq:decentralized_objective}
\end{align}

Intuitively, the first term optimizes each client’s local learning objective, while the second enforces \emph{latent-space consensus} across neighbors. 
The projection matrices ${\mathbf{P}_i}$ act as learnable operators that align client-specific feature spaces with a shared latent geometry to enable collaboration. Therefore, the regularizer in \eqref{eq:decentralized_objective} provides a surrogate coordination signal that guides local model training by penalizing deviations from shared latent representations.

\vspace{-0.1cm}

\section{Proposed Algorithm}
\label{sec:algorithm}

To solve the decentralized objective in \eqref{eq:decentralized_objective}, which can be decoupled across clients and classes at each step, we adopt an \emph{alternating optimization} strategy. 
At each communication round $t$, every client performs three main operations: 
(i) updates its local model parameters $\mathbf{w}_i$ guided by the latent-consensus regularizer, 
(ii) forms and exchanges compressed class-wise latent representations, and 
(iii) updates the projection (translator) matrix $\mathbf{P}_i$. 
Note that the gradient of the local loss with respect to $\mathbf{P}_i$ is zero, since the local task does not depend on the projection matrix.
{The latent-consensus regularizer is differentiable (or sub-differentiable) with respect to $\mathbf{w}_i$ through the tapped representation $\mathbf{v}_i(\mathbf{w}_i^{l_i};\xi)$, and gradients propagate through the network prefix via standard backpropagation.}

\begin{algorithm}[t]
\caption{\method: Multi-Modal Architecture-Agnostic Latent-Consensus Training}
\label{alg:comfed}
\begin{algorithmic}[1]
\Require Rounds $T$, graph $G$, learning rates $(\eta_w,\eta_P)$, regularizer weight $\lambda$, distance measure $\phi$
\For{$t=0$ to $T-1$}
  \For{each client $i$ in parallel}
    \State Update $\mathbf{w}_i^{(t+1)}$ using \eqref{eq:w_update_decent}
    \State For each $m\in\mathcal{M}$, compute $\widehat{\bar{\mathbf{v}}}_{i,m}^{(t)}$ and $\mathbf{u}_{i,m}^{(t)}=\mathbf{P}_i^{(t)}\widehat{\bar{\mathbf{v}}}_{i,m}^{(t)}$
    \State Exchange $\{\mathbf{u}_{i,m}^{(t)}\}_{m}$ with neighbors $j\in\mathcal{N}_i$ (or upload to PS)
    \State Compute $\mathbf{u}^{\star,(t)}_{(i,m)}$ via \eqref{eq:optimal_rep} (or receive global $\mathbf{u}^{\star,(t)}_{(m)}$ from PS)
    \State Update $\mathbf{P}_i^{(t+1)}$ using \eqref{eq:P_update_decent}
  \EndFor
\EndFor
\end{algorithmic}
\end{algorithm}

The alternating procedure proceeds as follows.  
First, each client updates its local weights {by
\begin{align}
\nonumber \mathbf{w}_i^{(t+1)}&\!=\!\mathbf{w}_i^{(t)}\!-\!\eta_w \nabla_{\mathbf{w}_i} \mathcal{L}_{\text{local}}^i\!\big(\mathbf{w}_i^{(t)}\big)\\
&-\!\frac{\eta_w\lambda}{|\mathcal{N}_i|}\sum_{j\in\mathcal{N}_i}
\nabla_{\mathbf{w}_i}\phi\!\left(\mathbf{P}_i\,\mathbf{v}_i(\mathbf{w}_i^{l_i}),\ \mathbf{u}_{j}\right),
\label{eq:w_update_decent}
\end{align}
}
where the round $t$ received neighbor representations $\{\mathbf{u}_{j}\}_{j\in\mathcal{N}_i}$ are treated as constants for $\mathbf{w}_i^{(t+1)}$ update in \eqref{eq:w_update_decent}.
The latent-consensus regularizer promotes alignment between local representations and those of neighboring clients. Therefore, each client uses its local training loss together with its representation regularizer to guide its model update.

Next, each client forms its compressed class-wise latent representations as
\(\widehat{\mathbf{u}}_{i,m}^{(t)}=\mathbf{P}_i^{(t)}\,\widehat{\bar{\mathbf{v}}}_{i,m}^{(t)}\), and exchanges the set $\{\widehat{\mathbf{u}}_{i,m}^{(t)}\}_{m\in\mathcal{M}}$ with its neighbors $j\in\mathcal{N}_i$. 
This communication step involves transmitting only one $d$-dimensional vector per class and per client, {allowing for} substantial communication savings compared to parameter-level exchanges.

After receiving neighbor representations, each client updates its projection matrix by pulling its projected class means toward the corresponding targets:
\begin{align}
\mathbf{P}_i^{(t+1)} =
\mathbf{P}_i^{(t)}-\eta_P\,\lambda
\frac{1}{|\mathcal{N}_i|}\sum_{j\in\mathcal{N}_i}
\nabla_{\mathbf{P}_i}\,\phi\!\left(\mathbf{P}_i\,\mathbf{v}_i(\mathbf{w}_i^{l_i}),\ \mathbf{u}_{j}\right).
\label{eq:P_update_decent}
\end{align}

For each class $m$, the regularization term involves multiple neighbor pulls
\begin{align}
\sum_{j\in\mathcal{N}_i}
\phi\Big(\mathbf{P}_i^{(t)}\,\widehat{\bar{\mathbf{v}}}_{i,m}^{(t)},\ 
\widehat{\mathbf{u}}_{j,m}^{(t)}\Big),
\label{eq:multiple_pulls}
\end{align}
where $\widehat{\mathbf{u}}_{j,m}^{(t)}=\mathbf{P}_j^{(t)}\,\widehat{\bar{\mathbf{v}}}_{j,m}^{(t)}$.
Assuming $\phi(\cdot,\cdot)$ is convex in its first argument and admits (sub)gradients, the multiple pulls can be replaced by a single per-class target $\mathbf{u}^{\star,(t)}_{(i,m)}$ obtained as
\begin{align}
\min_{[\widehat{\mathbf{u}}_{i,m}^{(t)}]}  
\sum_{j\in\mathcal{N}_i\cup\{i\}}
\phi\left(\widehat{\mathbf{u}}_{i,m}^{(t)},\ \widehat{\mathbf{u}}_{j,m}^{(t)}\right),
\label{eq:optimal_rep}
\end{align}
leading to the simplified projection update
\begin{align}
\min_{\mathbf{P}_i^{(t)}} 
\phi\Big(\mathbf{P}_i^{(t)}\,\widehat{\bar{\mathbf{v}}}_{i,m}^{(t)},\ 
\mathbf{u}^{\star,(t)}_{(i,m)}\Big).
\label{eq:optimal_rep2}
\end{align}

Two practical instances of $\phi$ are considered.  
For $\phi(x,y)=\|x-y\|_2^2$, the consensus target or in other words, the optimal solutions to \eqref{eq:optimal_rep}, correspond to the per-class arithmetic mean.  
For $\phi(x,y)=\|x-y\|_2$, it corresponds to the geometric median, yielding robustness to outliers. 
Both distance metrics admit closed-form (sub)gradients with respect to $\mathbf{P}_i$.

{In \eqref{eq:optimal_rep}, although the optimal representations at the current round admit a closed-form solution, we are deliberately only interested in taking a single projection matrix update step towards them, since the projection matrices and model weights are learned concurrently. Therefore, we can replace the multiple pulls in \eqref{eq:multiple_pulls} with a single pull in \eqref{eq:optimal_rep2} given that $\phi$ is convex.}

{In a (PS-based) setting, the server is the center of a star-based communication graph topology, but logically it can act as a communication hub that induces a logical fully connected graph, such that all clients share the same per-class consensus targets. Consequently, moving from the decentralized formulation, and using the single pull trick deduced in \eqref{eq:multiple_pulls}, \eqref{eq:optimal_rep}, and \eqref{eq:optimal_rep2}
. Clients can upload their class-wise latent representations $\{\widehat{\mathbf{u}}_{i,m}^{(t)}\}_{m\in\mathcal{M}}$ to the server, enabling the server to compute a global per-class representation and broadcast it back to all clients. As a result, the local update rules and optimization objective remain unchanged; only the entity computing the per-class representation consensus differs.}

{From a privacy perspective, the projection matrices $\{\mathbf{P}_i\}$ align client-specific feature spaces into a shared latent geometry while ensuring that only low-dimensional, class-wise projected representations are exchanged. As a result, the communicated information lacks direct semantic meaning and is difficult to interpret or invert. Although formal privacy guarantees are not the main focus of this work, operating in a low-dimensional latent space naturally facilitates the integration of differential privacy mechanisms by allowing calibrated noise to be added to the shared latent statistics.}

{\textbf{Uplink Payload and Overhead.}
The communication bottleneck in FL lies in the uplink transmission. In \ours{}, each client transmits only class-wise projected latent representations of dimension $d$, resulting in a per-round uplink payload of order $\mathcal{O}(|\mathcal{M}|\cdot d)$, independent of, and much smaller than, the local model size. The additional client-side memory consists of storing a single projection matrix, and the extra computation is limited to lightweight projection and averaging operations. Aggregation is performed in the same low-dimensional latent space, leading to minimal server overhead and low latency.
}

\begin{table}[t]
  \centering
  \caption{Cumulative communication overhead required to first achieve each target accuracy (USC dataset). ``NA'' denotes that the accuracy threshold was not attained.}
  \label{tab:performance}
  \small
  \setlength{\tabcolsep}{3pt}
  \begin{tabular}{@{}l c c@{}}
    \toprule
    \textbf{Algorithm} & \textbf{Uplink Comp.} & \textbf{Comm. Cost (40 / 50 / 58\%)} \\
    \midrule
    \ours{} 
      & $\mathcal{O}(|\mathcal{M}|d)$ 
      & \textbf{269 KB} / \textbf{410 KB} / \textbf{768 KB} \\
    Harmony (UniFL) 
      & $\mathcal{O}(|\mathbf{w}|)$ 
      & 4.39 GB / 9.57 GB / NA \\
    FedMD 
      & $\mathcal{O}(|\mathcal{D}_{\text{public}}|\!\cdot\!|\mathcal{M}|)$ 
      & 723 KB / 1.73 MB / NA \\
    FedIoT 
      & $\mathcal{O}(|\mathbf{w^{g}}|)$ 
      & NA / NA / NA \\
    \bottomrule
  \end{tabular}
\end{table}

\begin{figure}[t]
  \centering
  \includegraphics[scale=0.39]{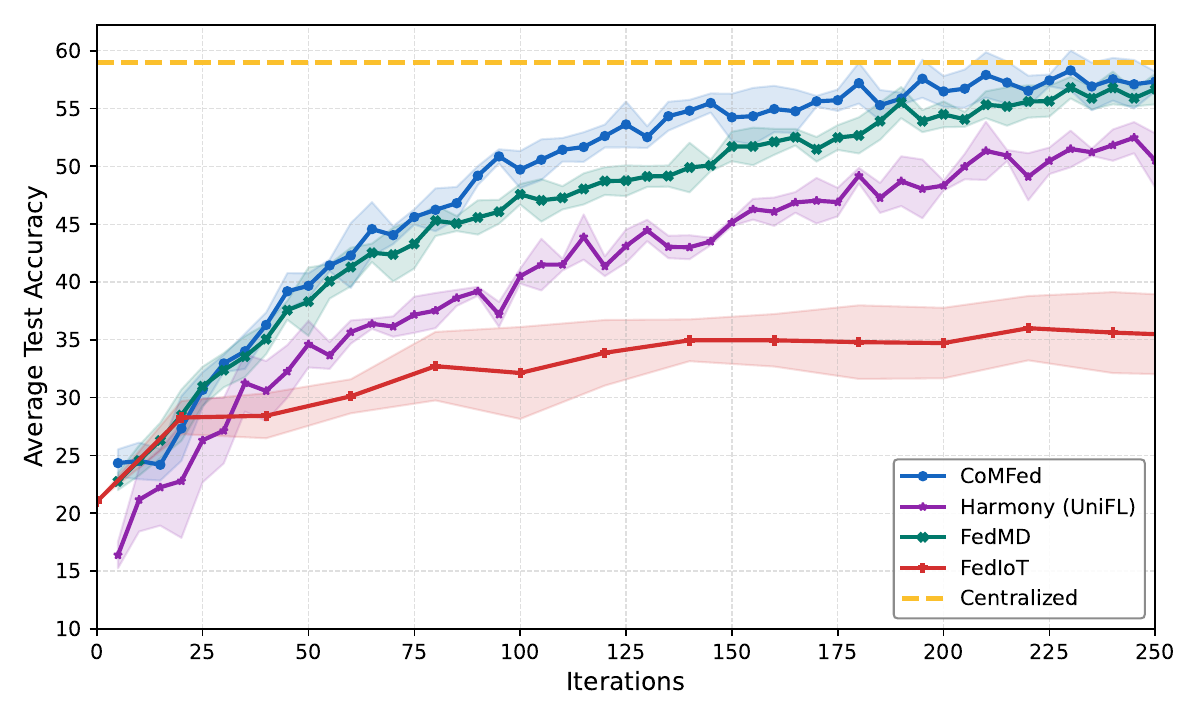}
 \caption{Average test accuracy over training iterations (USC dataset)}

  \label{fig:activity_avg_acc_all}
\end{figure}

\begin{figure}[t]
  \centering
  \includegraphics[scale=0.39]{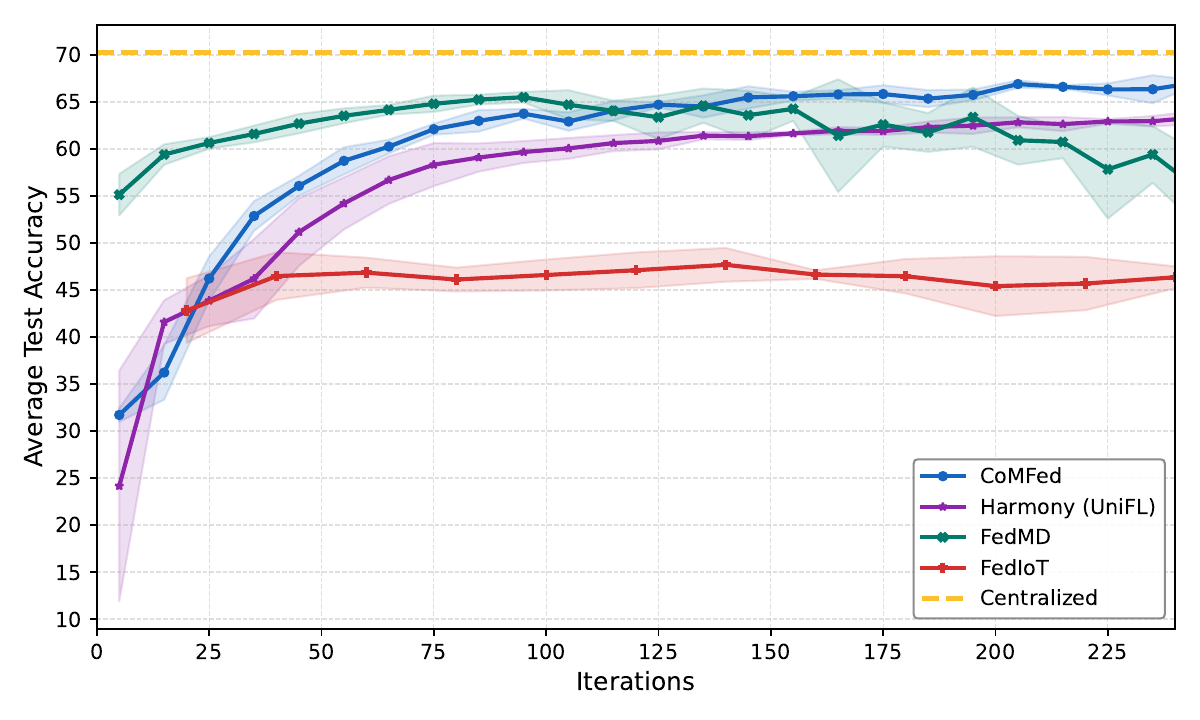}
\caption{Average test accuracy over training iterations (DeepSense blockage-severity dataset)}
  \label{fig:DeepSense_avg_acc_all}
\end{figure}
\section{Numerical Results}\label{sec:num}
{We evaluate the proposed method on two real-world collected multi-modal datasets.
Experiments are first conducted on the USC-HAD activity recognition dataset \cite{zhang2012usc}, which contains inertial measurements collected from 14 wearable devices, each treated as a client. Although both accelerometer (ACC) and gyroscope (GYR) modalities are available during data collection, FL training is performed under heterogeneous conditions: 3 clients use ACC only, 3 use GYR only, and 8 use both modalities, following \cite{ouyang2023harmony}. Single-modality clients employ a lightweight CNN encoder followed by a GRU-based temporal classifier, while multi-modal clients process ACC and GYR through parallel CNN branches whose outputs are fused prior to temporal modeling. Latent features are extracted from a 128-dimensional representation layer, and the projection matrices $\{\mathbf{P}_i\}$ are randomly initialized.}

{To complement USC-HAD, we further evaluate \ours{} on the DeepSense blockage-duration prediction dataset \cite{charan2022vision}, which consists of real-world mmWave and LiDAR measurements collected across six deployment scenarios. The task is to classify blockage duration into four levels. Six clients are formed: two with mmWave only, two with LiDAR only, and two with both modalities. Single-modality clients use modality-specific CNN encoders, while multi-modal clients fuse LiDAR and mmWave representations prior to classification. Compared to USC-HAD, this dataset involves fewer classes, stronger modality disparity, and higher heterogeneity across clients, providing a complementary stress test.}

{We compare \ours{} against three multi-modal FL baselines with different methodological assumptions. Harmony \cite{ouyang2023harmony} performs modality-wise FL training by partitioning clients according to modality availability. FedIoT \cite{zhao2022multimodal} leverages unlabeled data at the clients to train modality-specific autoencoders, while a labeled public dataset is used at the server to train a global classifier on the encoded representations. We additionally include FedMD \cite{li2019fedmd}, a distillation-based heterogeneous FL approach in which clients first train locally on private data and subsequently exchange logits evaluated on a shared public dataset to align heterogeneous models through knowledge distillation on the public dataset. To enable comparison with FedIoT and FedMD, one client’s data is treated as public.}
{We also include a centralized benchmark where all data are pooled, and a single model is trained.}

{For fairness and comparability, all methods perform the same total number of model update iterations (250 for both datasets) using a learning rate of $10^{-3}$. For \ours{}, the projection matrices are trained jointly with the local models using the same learning rate, a regularization weight of $\lambda=0.4$, and 10 projection update steps per iteration. Results are averaged over five Monte Carlo runs.}

\begin{figure}[t]
  \centering
  \includegraphics[scale=0.39]{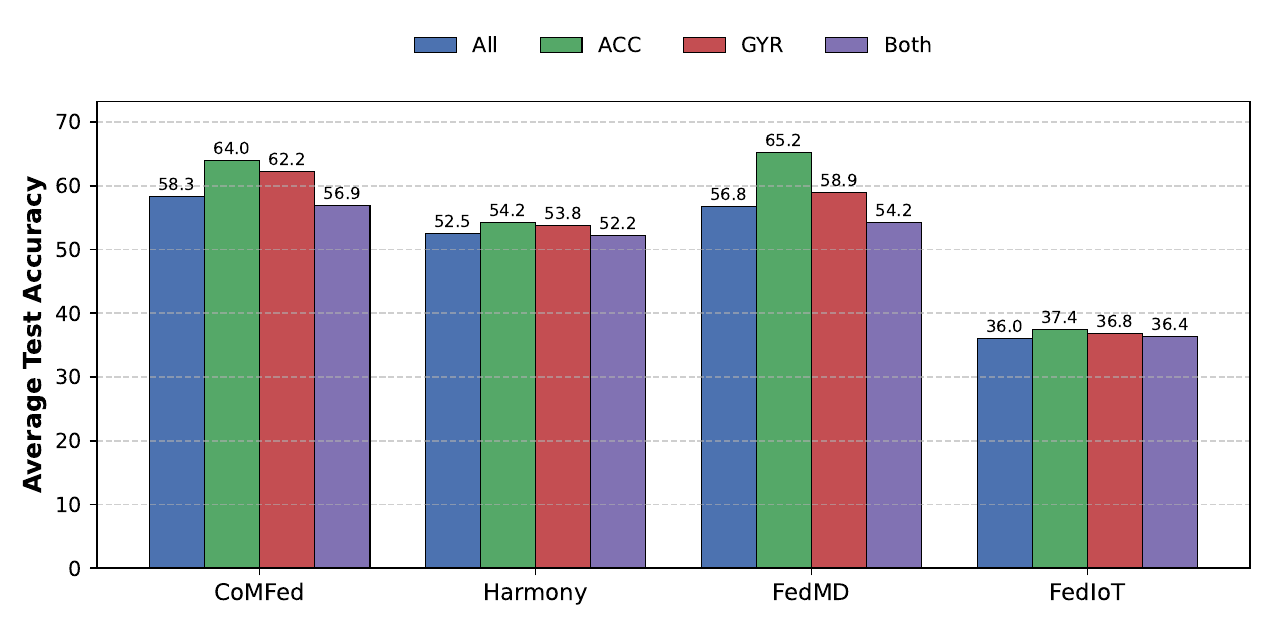}
  \caption{Accuracy comparison among algorithms and across modality subsets (USC-dataset).}
  \label{fig:activity_barplot}
\end{figure}

{Figs.~\ref{fig:activity_avg_acc_all} and \ref{fig:activity_barplot} show that \ours{} achieves faster convergence and higher average accuracy than the baselines. While FedMD performs competitively and can outperform \ours{} on specific modalities (e.g., ACC-only clients), \ours{} attains stronger average performance across the remaining modality subsets.}

\begin{figure}[t]
  \centering
  \includegraphics[scale=0.39]{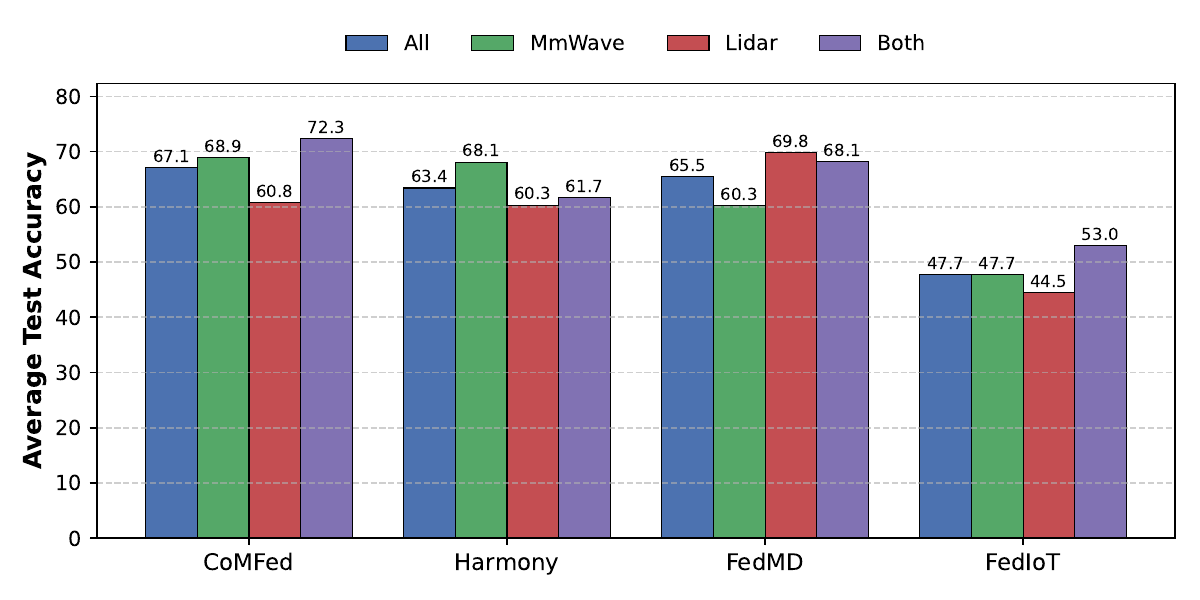}
  \caption{Accuracy comparison among algorithms and across modality subsets (DeepSense-dataset).}
  \label{fig:DeepSense_barplot}
\end{figure}

{On the DeepSense dataset (Figs.~\ref{fig:DeepSense_avg_acc_all} and \ref{fig:DeepSense_barplot}), FedMD achieves strong performance on LiDAR-only clients but exhibits noticeable fluctuations during training. This behavior arises from its reliance on public data knowledge distillation, which can produce inconsistent supervisory signals when the public dataset does not adequately reflect the heterogeneity of private client data. In contrast, \ours{} exhibits smoother convergence and higher average accuracy across all modality subsets by avoiding dependence on potentially unavailable or poorly matched public data and instead leveraging a robust latent-space consensus mechanism for collaboration.}

\begin{figure}[t]
  \centering
  \includegraphics[scale=0.39]{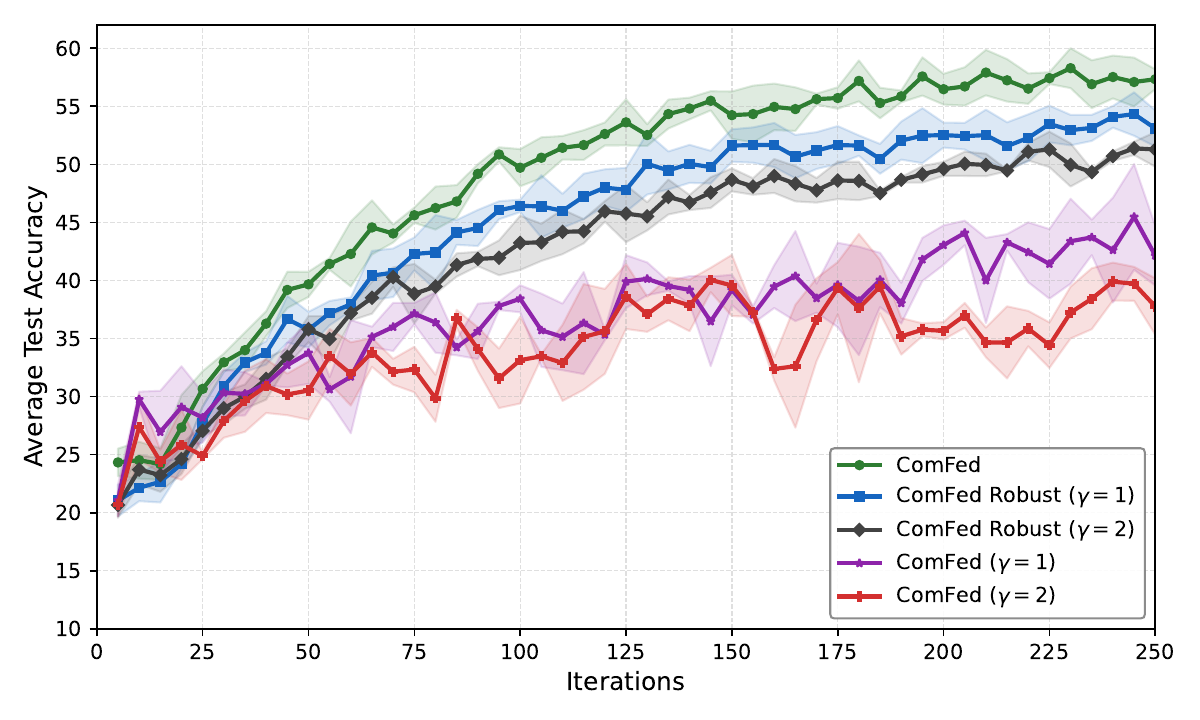}
  \caption{Robustness to attacks median vs mean methods for USC-dataset ($\gamma$ denotes the number of {Byzantine} clients).}
  \label{fig:robustness}
\end{figure}

Robustness results in Fig.~\ref{fig:robustness} show that the geometric-median consensus variant of \ours{} maintains higher accuracy as the number of Byzantine clients increases, demonstrating resilience to corrupted or noisy representations.

\begin{figure}[t]
  \centering
  \includegraphics[scale=0.39]{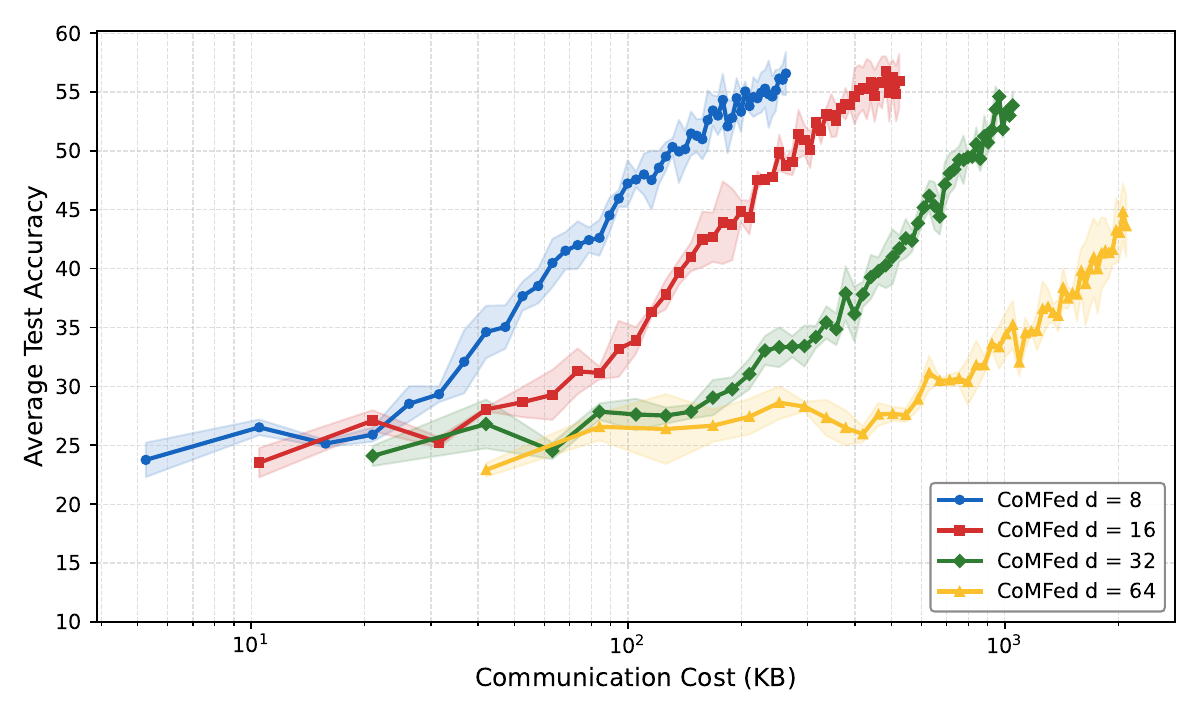}
  \caption{Average accuracy vs communication cost for different projection dimensions (USC-dataset).}
  \label{fig:cost_projections}
\end{figure}

{Table~\ref{tab:performance} summarizes the uplink communication complexity and cumulative communication overhead required to reach target accuracies. In summary, the lower communication overhead needed, and the non-reliance on a high-quality public dataset, show the potential advantages of {\ours} over different baselines. Finally, Fig.~\ref{fig:cost_projections} shows that increasing the projection dimension $d$ improves accuracy up to a point before degradation occurs, while the cost scales linearly with $d$.}

\vspace{-0.1cm}


\vspace{-0.1cm}
\section{Conclusion}\label{sec:conclusion}
We proposed a communication-efficient multi-modal FL framework that aligns heterogeneous client representations through low-dimensional projections, using class-wise latent statistics as the core collaboration signal. The method enables clients with diverse architectures and modalities to jointly learn both their local models and a shared latent space, while operating over arbitrary communication topologies.
\vspace{-0.1cm}
\bibliographystyle{IEEEtran}
\bibliography{references} 

\end{document}